\documentclass[10pt,twocolumn,letterpaper]{article}

\usepackage{3dv}
\usepackage{times}
\usepackage{epsfig}
\usepackage{graphicx}
\usepackage{amsmath}
\usepackage{amssymb}

\usepackage{float}
\usepackage{booktabs}
\usepackage{multirow}
\usepackage{cite}
\usepackage{stfloats}
\usepackage{amsmath}
\usepackage{bbding} 
\usepackage[normalem]{ulem} 

\usepackage{paralist}
\usepackage{xcolor}

\usepackage[pagebackref=true,breaklinks=true,colorlinks,bookmarks=false]{hyperref}

\threedvfinalcopy

\def\threedvPaperID{39}

\begin{document}

\title{Efficient Human Pose Estimation via 3D Event Point Cloud}
\author{Jiaan Chen$^{1,}$\thanks{The first two authors contribute equally to this work.}\, ,
~~Hao Shi$^{1,*}$,
~~Yaozu Ye$^1$,
~~Kailun Yang$^2$,
~~Lei Sun$^{1,3}$,
~~Kaiwei Wang$^1$\\
\normalsize
$^1$Zhejiang University
\normalsize
~$^2$Karlsruhe Institute of Technology
\normalsize
~$^3$ETH Z\"urich\\
{\tt\small \{chenjiaan,haoshi,yaozuye,leo\_sun,wangkaiwei\}@zju.edu.cn, kailun.yang@kit.edu}
}

\maketitle

\begin{abstract}
\vspace{-0.75em}
Human Pose Estimation (HPE) based on RGB images has experienced a rapid development benefiting from deep learning. However, event-based HPE has not been fully studied, which remains great potential for applications in extreme scenes and efficiency-critical conditions. In this paper, we are the first to estimate 2D human pose directly from 3D event point cloud. We propose a novel representation of events, the rasterized event point cloud, aggregating events on the same position of a small time slice. It maintains the 3D features from multiple statistical cues and significantly reduces memory consumption and computation complexity, proved to be efficient in our work. We then leverage the rasterized event point cloud as input to three different backbones, PointNet, DGCNN, and Point Transformer, with two linear layer decoders to predict the location of human keypoints. We find that based on our method, PointNet achieves promising results with much faster speed, whereas Point Transfomer reaches much higher accuracy, even close to previous event-frame-based methods. A comprehensive set of results demonstrates that our proposed method is consistently effective for these 3D backbone models in event-driven human pose estimation. Our method based on PointNet with $2048$ points input achieves $82.46mm$ in MPJPE$_{3D}$ on the DHP19 dataset, while only has a latency of $12.29ms$ on an NVIDIA Jetson Xavier NX edge computing platform, which is ideally suitable for real-time detection with event cameras. Code is available at \url{https://github.com/MasterHow/EventPointPose}.
\end{abstract}

\vspace{-1.75em}
\section{Introduction}
\vspace{-0.25em}
Human Pose Estimation (HPE) aims to predict the keypoints of each person from perceived signals. The predicted results could be used for understanding people in images and videos, with wide potential applications such as action recognition~\cite{yan2018stgcn, shi2019skelaction} and animations~\cite{willett2020pose2pose}.
Predicting 2D human poses~\cite{cheng2020higherhr, xiao2018simbase,fang2017rmpe,li2021posetrans} from RGB images and recovering 3D poses~\cite{Chen_2017_CVPR3d2dmatch, iskakov2019learnable3d} have developed rapidly in the last years.

\begin{figure}[!t]
\centering
\includegraphics[width=0.8\linewidth]{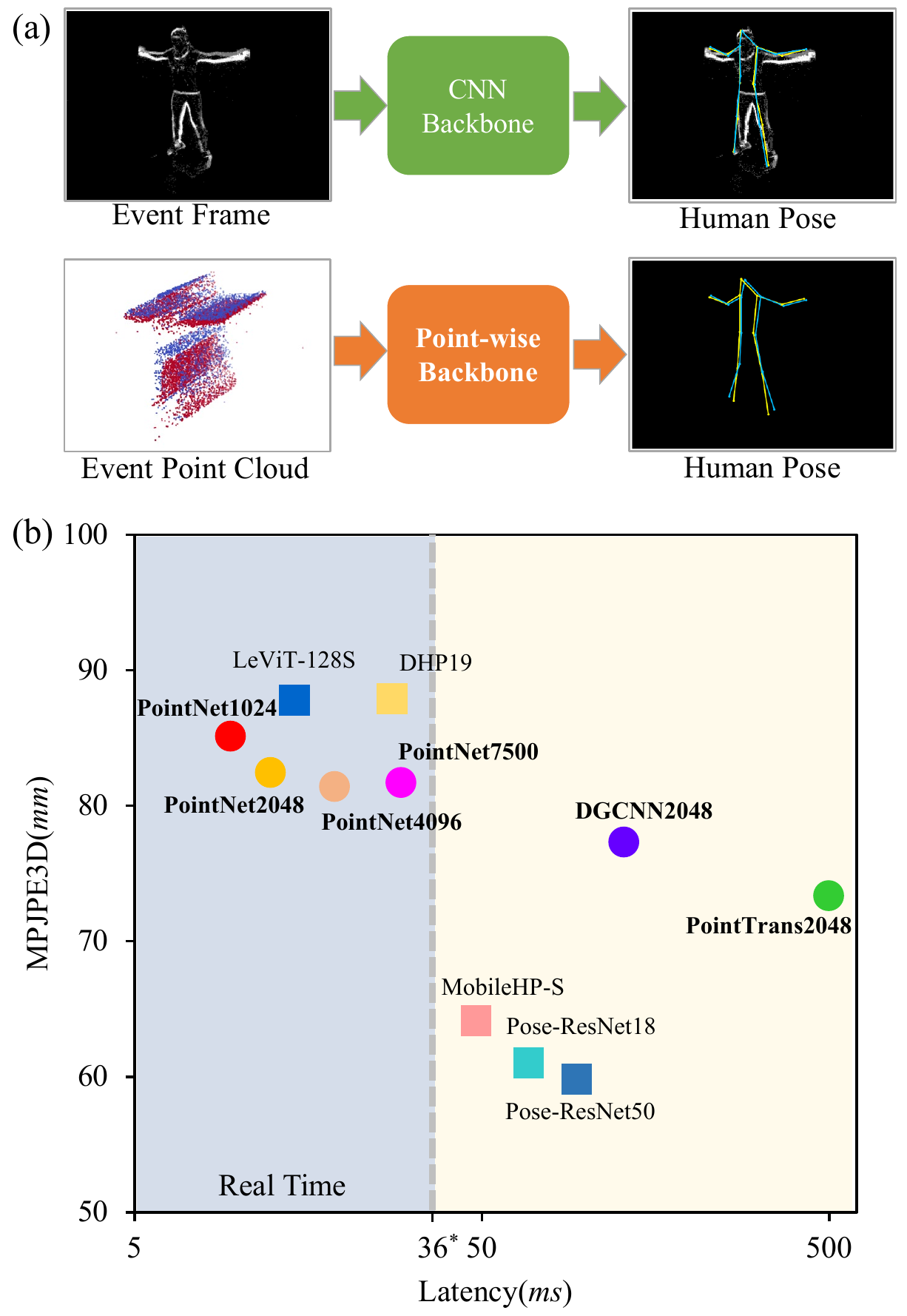}
\vskip-1ex
\caption{(a) 2D event frame based human pose estimation paradigm \textit{vs.} the proposed novel 3D event point cloud based paradigm, (b) Latency \textit{vs.} Mean Per Joint Position Error with a logarithmic x-axis of 2D CNN backbones (square markers) and our 3D proposed pipeline (circular markers). *The real-time criterion is statistically obtained on the DHP19 test dataset~\cite{calabrese2019dhp19}.
}
\label{intro}
\vskip-3.5ex
\end{figure}

Yet, HPE still meets challenges with the drawbacks of frame-based cameras. The prediction of keypoints in scenarios with motion blur or high dynamic range will be inaccurate. Event cameras, such as the Dynamic Vision Sensor (DVS)~\cite{gallego2020eventsurvey}, with higher dynamic range and larger temporal resolution, can tackle these disadvantages, which could maintain stable output in such extreme scenes.
Enrico~\textit{et al.}~\cite{calabrese2019dhp19} first addressed the application of DVS cameras for HPE to promote the development of efficient HPE, and introduced the first DVS dataset for 3D human pose estimation.
Due to the high performance of 2D heatmap representation in 2D HPE, a direct way to explore HPE with DVS is to process the events as an event frame or event voxels with Convolutional Neural Networks (CNNs) and predict 2D heatmaps to yield pose estimation~\cite{calabrese2019dhp19, scarpellini2021lifting}.
Still, the CNN-based methods do not sufficiently consider the asynchronous sparse characteristics of this data format, and the high amount of computation brings high latency, which is limited in real-time processing for DVS (see Fig.~\ref{intro}).

However, owing to the distinctiveness of events for its multidimensional nature, we aim to explore a new estimation paradigm to unleash the potential of the device, with the characteristic of sparse and low-latency output, for HPE.
In 3D point cloud processing, classification~\cite{qi2017pointnet, qi2017pointnet++} and segmentation~\cite{wang2019dgcnn, zhao2021pointtrans} of point clouds have been explored via deep learning.
Further, 3D HPE through 3D point clouds has also been studied~\cite{zhou2020learning3Dpoint, chan20143pointcloudsystem}.
Nevertheless, predicting human keypoints directly through event streams is scarcely addressed in the state of the art.
To fully unfold the potential of efficient HPE with DVS by processing events from event point sets, we treat events as multidimensional data and predict the 2D joints for a single event camera via backbones established for point cloud processing.
To our best knowledge, we are the first to explore 2D HPE directly from 3D event point clouds (Fig.~\ref{intro}).
Instead of utilizing the raw event data, we propose a rasterized format of event point cloud, aggregating events on the same position of a small time slice to maintain the multidimensional features of events.
We then leverage the rasterized event point cloud as input to a 3D learning backbone and decode the output of the backbone.
Three different point cloud learning backbones, PointNet~\cite{qi2017pointnet}, DGCNN~\cite{wang2019dgcnn}, and Point Transformer~\cite{zhao2021pointtrans} are employed in our framework, with two linear layer decoders to predict the location of human points. Finally, the 2D predictions are transformed to 3D via triangulation.

Following this strategy, a comprehensive variety of experiments is performed on the DHP19 dataset~\cite{calabrese2019dhp19}. We further test the HPE models with our own DAVIS camera, which achieve robust results on new, previously unseen domains. 
Our approach based on PointNet exceeds the method of event-frame-based DHP19~\cite{calabrese2019dhp19} in MPJPE$_{3D}$ and a lower latency is reserved, $9.43ms$ for $1024$ points as input.

On a glance, we deliver the following contributions:
\begin{compactitem}
  \item We explore the feasibility of estimating human pose from 3D event point clouds directly, which is the first work from this perspective to our best knowledge.
  \item We demonstrate the effectiveness of well-known LiDAR point cloud learning backbones for event point cloud based human pose estimation.
  \item We propose a new event representation--\emph{rasterized event point cloud}, which maintains the 3D features from multiple statistical cues and significantly reduces memory consumption and computational overhead with the same precision.
  \item We explore labeling methods for the event point cloud dataset and show that the \emph{Mean Label} is more effective than the \emph{Last Label}, considering that the temporal resolution of labels is usually smaller than that of events.
\end{compactitem}

\section{Related work}
\vskip-0.5ex
\subsection{Human pose estimation}
\vskip-0.5ex
In recent years, human pose estimation, which usually includes 2D and 3D HPE, has been an important part of computer vision community, and most of the previous researches are investigated based on RGB frames.
In the field of 2D single-person pose estimation, recent mainstream CNN-based methods usually predict heatmaps for joints and select the positions with maximum confidence as the final prediction~\cite{newell2016stackedhourglass, wei2016convposemachain, xiao2018simbase} and these kinds of solutions usually outperform counterparts which predict keypoints coordinates directly~\cite{toshev2014deeppose, Carreira_errorfeedback}.
Another group of efforts attempts to investigate HPE with integral regression methods and learnable operations~\cite{sun2018integral}.
Top-down~\cite{fang2017rmpe, xiao2018simbase} and bottom-up~\cite{cao2017openpose, GengSXZW21DEKR, cheng2020higherhr} are two study perspectives for multi-person pose estimation. 
2D HPE could be developed on several datasets, such as MPII~\cite{mpii}, MSCOCO~\cite{lin2014microsoftcoco}, PoseTrack~\cite{andriluka2018posetrack}, CrowdPose~\cite{li2019crowdpose}, FLIC~\cite{FLIC}, and LSP~\cite{LSP}, \textit{etc.} 

Existing approaches reconstruct the 3D pose from single~\cite{li20143dmono, Chen_2017_CVPR3d2dmatch, Martinez_2017_ICCV3dbase} or multiple~\cite{iskakov2019learnable3d} camera views.
Multi-view approaches ease the problem of occlusion and ambiguities and achieve higher accuracy than single-view methods.
Datasets such as HumanEva~\cite{sigal2010humaneva}, Human3.6M~\cite{ionescu2013human36m}, CMU Panoptic~\cite{Joo_2017_TPAMI}, and MPI-INF-3DHP~\cite{mono-3dhp2017} are usually used for 3D researches.
For real-time applications, knowledge distillation could be used to obtain small models~\cite{zhang2019fasthpe} and lightweight frameworks for mobile devices could achieve fast speeds~\cite{choi2021mobilehumanpose}. We do not apply any model compression techniques, and a fair comparison is kept when exploring the new event point cloud based paradigm.
Our method falls into the category of multi-view approaches for single-person HPE, while predicting 2D results directly from 3D event point cloud, which is different from existing works.

\begin{figure*}[!t]
\centering
\includegraphics[width=1.0\linewidth]{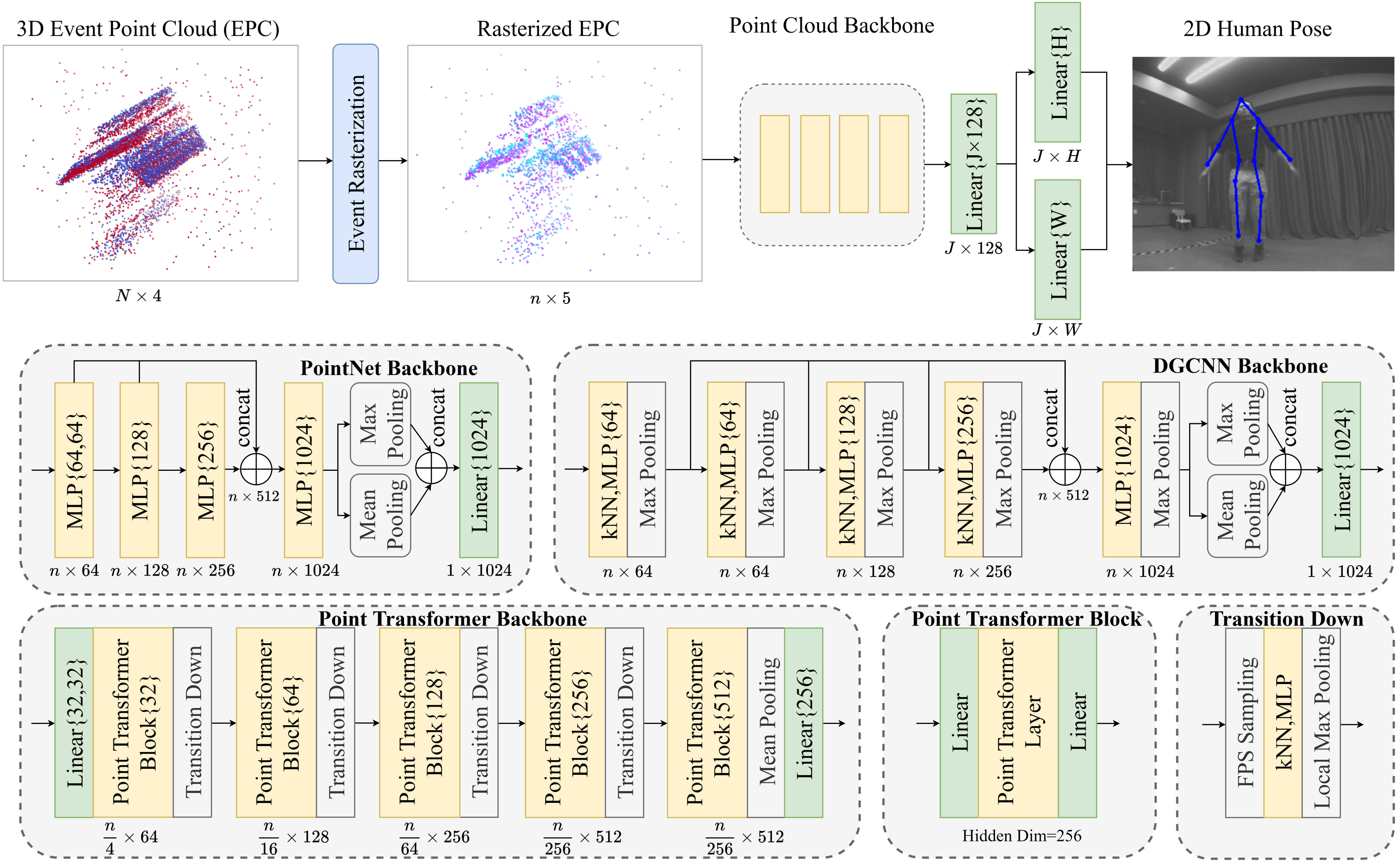}
\vskip-1ex
\caption{The proposed pipeline. The raw 3D event point cloud is first rasterized, and then processed by the point cloud backbone. The features output from the backbone are then connected to linear layers to predict two vectors. 2D positions of human keypoints are proposed via decoding the two vectors.}
\label{backbone}
\vskip-4ex
\end{figure*}

\vskip-0.5ex
\subsection{Event based vision}
\vskip-0.5ex
Event camera~\cite{gallego2020eventsurvey} is a kind of bio-inspired asynchronous sensor responding to changes in brightness on each pixel independently.
With high temporal resolution (in the order of $\mu s$) and high dynamic range (over $100dB$), it has huge potential in many extreme scenes, and has been proved effective in many applications such as deblurring~\cite{sun2021mefnet},
optical flow estimation~\cite{gallego2018unifying_contrast_event,gehrig2021eraft_optical_flow,pan2020single_optical_flow_event},
corner tracking~\cite{alzugaray2019asynchronous_tracking},
scene segmentation~\cite{alonso2019ev_segnet,zhang2020issafe,zhang2021exploring_event_segmentation},
visual odometry~\cite{zhou2021event_visual_odometry},
human motion capture and shape estimation~\cite{xu2020eventcap, zou2021eventhpe}.
To leverage the advantages of asynchronous events, different types of events representation are explored for event-based vision~\cite{gehrig2019endlearnrepre, maqueda2018eventframe, zhu2019unsupervised, sun2021mefnet, baldwin2022TORE}.
Most algorithms manage events to event frames to adapt CNN-based methods~\cite{maqueda2018eventframe, zhu2018evflow}.

Responding only to dynamic objects brings fixed-position event camera native advantages in human action recognition, gesture recognition, and gaits identification~\cite{wang2019space, baby2017dynamichumanactivity, amir2017lowsnn, wu2020multipathhumanaction, wang2021eventgait}.
Event-based HPE also needs to be explored to deal with extreme scenes and efficiency-critical conditions. Enrico~\textit{et al.}~\cite{calabrese2019dhp19} introduce the first DVS dataset for 3D human pose estimation with event streams, and it is the only event-based human pose dataset recorded in the real world till now. 
They propose a lightweight but effective CNN model to predict human pose from constant count event frames and obtain 3D results via triangulation.
Because of the limited quantity of real event datasets, generating events in simulators becomes an alternative~\cite{gehrig2020video_to_events, rebecq2018esim_simulator,hu2021v2e}.
Alex~\textit{et al.}~\cite{zhu2021eventgan} and Scarpellini~\textit{et al.}~\cite{scarpellini2021lifting} test their methods on the simulated datasets and Scarpellini~\textit{et al.}~\cite{scarpellini2021lifting} propose a monocular method to predict 3D coordinates. Baldwin~\textit{et al.}~\cite{baldwin2022TORE} attain a competitive 3D result on the DHP19 dataset via a new representation of events and a process of a 3D estimation model which enhances temporal consistency and enables multi-camera feature fusion.
\cite{scarpellini2021lifting} introduces a monocular method and \cite{baldwin2022TORE} relies on a temporal process, and both do not fit in a fair comparison with our method.

Further, most of these works are mainly conducted with 2D CNNs, while the natural characteristic of events as asynchronous signals with high temporal resolution is undermined. Exploiting the advantage of events through raw event points is also studied in human action recognition~\cite{wang2019space, chen2020eventdgcnn}. They use LiDAR point clouds processing neural networks to process events and achieve great action classification results.
Different from the action classification, human pose estimation via regressing keypoints coordinates, has still not been explored through event point clouds, which is the main contribution of our work. We explore and prove the efficiency of event-based HPE through processing the event point clouds.

\section{Methodology}
The pipeline of our proposed method is illustrated in Fig.~\ref{backbone}. The raw 3D event point cloud is first rasterized, and then processed by the point cloud backbone, connected with two linear layers to predict 2D positions of human keypoints. We then describe the methods in detail.
\subsection{Dataset}
\label{sec:approach_dataset}
We mainly explore human pose estimation based on the DHP19 dataset~\cite{calabrese2019dhp19}, which is the only dataset with real events for event-based HPE, and it is recorded by four event cameras.
In the previous event-frame-based work~\cite{calabrese2019dhp19}, they generated event frames for the four cameras, which are time synchronized, by accumulating $30k$ events for the four views in total.
To obtain the raw event point sets used for our approach, we follow the process of denoising and filtering in DHP19, but we do not accumulate constant count frames, instead we save the raw event point sets, in order to process through point-wise methods. For each event point, it is represented by $e = (x, y, t, p)$, where $(x,y)$ is the pixel location of the event, $t$ represents the timestamp in microsecond resolution, and $p$ is the polarity of the brightness change, $0$ for decrease and $1$ for increase. However, when we accumulate the events together, some cameras may not be calculated, leading to an empty point cloud or have too few points.
It also occurs when generating event frames which correspond to an empty image.
We remove the data with points fewer than $1024$ when training our model.

Same as the raw DHP19, we handle its four views together to generate the shared \emph{Mean Label}.
In DHP19, the streams of events from the four cameras are merged and have an ordered monotonic timestamp, which is convenient for processing data together.
When $N=30k$ events from the four views are accumulated in total, given by $E$ in Eq.~\ref{eq:1}, one 3D label, $gt_{mean}$ is produced.
The label is the mean value of the 3D coordinates for each joint generated in the window of $N$ events, followed by Eq.~\ref{eq:2}, and is shared by the four cameras.
\begin{equation}\label{eq:1}
E=\left\{E_{i} \mid i=1,2, \ldots N\right\},
\end{equation}
\begin{equation}\label{eq:2}
gt_{\text{mean}}=\operatorname{Mean}\left(gt_{T_{\min}}, gt_{T_{\min} +dt}, \ldots gt_{T_{\max}}\right),
\end{equation}
\begin{equation}
T_{\min } = \left\{\mathop{\arg\min}\limits_{T}\left(T-E_{1}(t)\right) \mid T \geq E_{1}(t) \right\}, 
\end{equation}
\begin{equation}
T_{\max } = \left\{\mathop{\arg\min}\limits_{T}\left(E_{N}(t)-T\right) \mid T \leq E_{N}(t)\right\}.
\end{equation}
After obtaining the 3D joints labels, we project them to 2D labels for a single camera view with the projection matrices.
We train the human pose estimation model under the dataset with 2D labels for each camera. Main experiments are designed in the \emph{Mean Label} setting. We also explore another label generation method, naming \emph{Last Label}, by assigning labels nearest to the last timestamp, and the results of the two kinds of dataset setting will be discussed in Sec.~\ref{sec:ablation_studies}. 

\subsection{Data preprocess}
\begin{figure*}[!t]
\centering
\includegraphics[width=0.86125\linewidth]{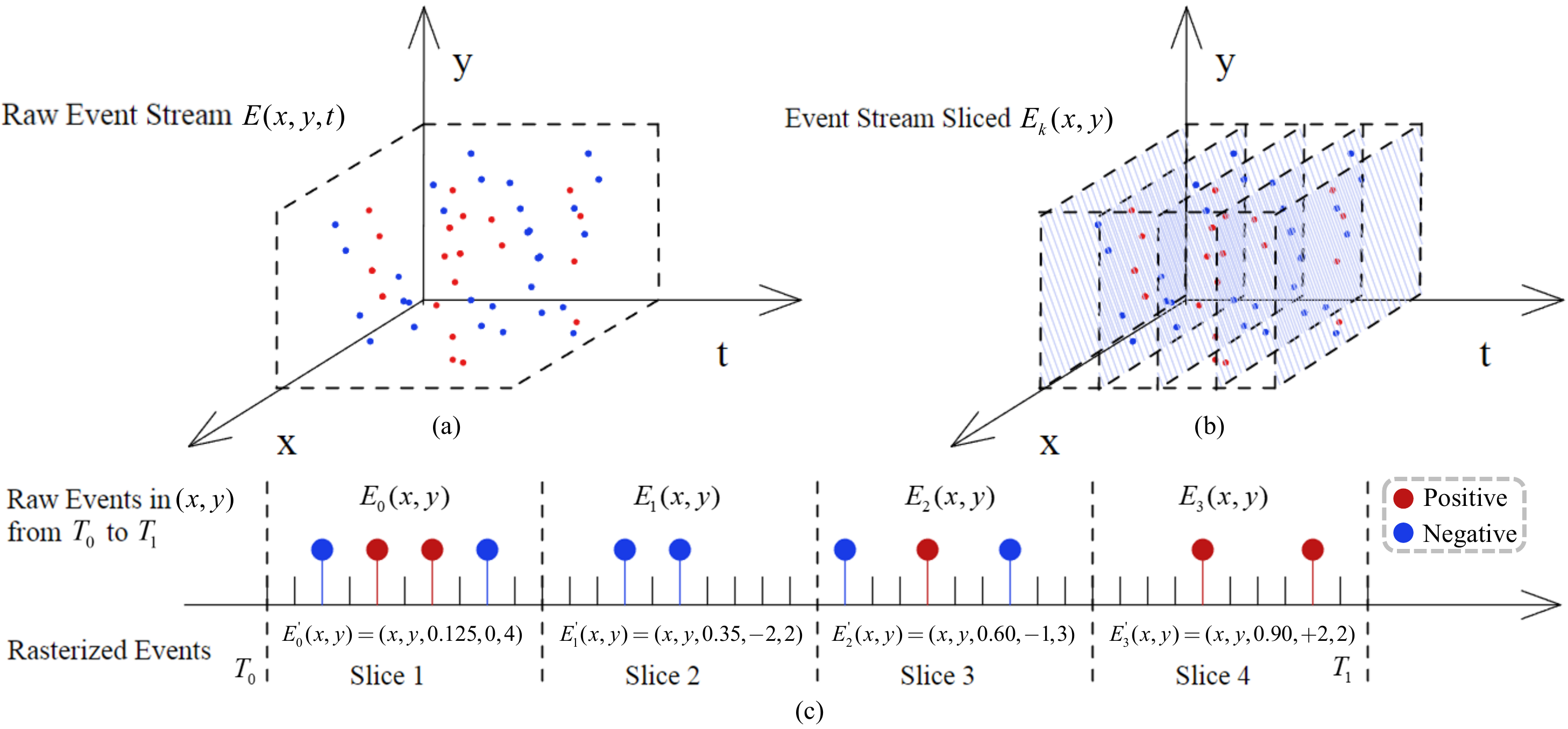}
\vskip-1ex
\caption{Schematic diagram of event point cloud rasterization. (a) Raw 3D event point cloud input, (b) Time slice of event point cloud, (c)~Rasterized event point cloud at $(x, y)$ position. Note that the rasterization process preserves the discrete nature of the point cloud, rather than the 2D image.}
\label{ras}
\vskip-3.5ex
\end{figure*}

\noindent\textbf{Rasterized point cloud format of the event.}
The event camera generates events with microsecond temporal resolution once a pixel brightness change exceeds a threshold.
Once all events are fed into the network, the forward propagation will be quite slow, and the memory consumption is also an issue to be considered.
We expect a way to retain as much information as possible while significantly reducing the number of events.
Here, we propose an event rasterization method. For events between $T_i$ and $T_{i+1}$, first, we split them into $K$ time slices.
Within each small time slice, we condense all events occurring on the same pixel into a new event.
Specifically, given all $M$ events on position $(x, y)$ in time slice $k$, where $p_{i}$ is converted to ${-}1$ for brightness decrease: 
\begin{equation}
E_k(x, y)=(x, y, t_i, p_i),\quad i=1,.., M,
\end{equation}
we use the following equations to obtain the rasterized event $E^\prime_k(x,y)$:
\begin{equation}
E^\prime_k(x,y)=(x, y, t_{avg}, p_{acc}, e_{cnt})
\end{equation}
\begin{equation}
t_{avg} = \frac{1}{M}\sum_i^Mt_i,\quad p_{acc} = \sum_i^Mp_i,\quad e_{cnt}=M,
\end{equation}
With this rasterization, we can reduce the number of events while retaining a considerable amount of event information, and this operation can be also seen as an aggregation of events in the time dimension.
We select $K{=}4$ in this work, which is appropriate to maintain the time resolution. And $t_{avg}$ in all $K$ slices are normalized to the range of $[0, 1]$. An example of the proposed event point cloud rasterization is illustrated in Fig.~\ref{ras} and the result is visualized in Fig.~\ref{backbone}, where the color represents the value of $p_{acc}$ and the point size for $e_{cnt}$.

\noindent\textbf{Sample points.}
As mentioned above, event points after being rasterized, are reduced via a kind of sampling, but retain enough event information.
Yet, the number of rasterized points is not equal for each event point cloud.
Therefore, we need to sample the data to the same numbers.
As a preprocess in this task, we consider the random point sampling method better than the furthest point sampling method, mainly considering the speed, where random point sampling is much faster than the furthest point sampling, retaining the advantage of real-time prediction.
The number of points input to the model is the main reason impacting the speed of the model, while also affecting the accuracy.
In Sec.~\ref{sec:ablation_studies}, we further explore the effect of the numbers after sampling on the trade-off between efficiency and accuracy in human pose estimation.

\subsection{Label representation}
By projecting 3D labels to 2D labels for a single camera view with the projection matrices, we have 2D labels shape as $(J, 2)$, where $J$ is the number of joints, $13$ for the DHP19 dataset, and $(x^{\prime}, y^{\prime})$ corresponds to pixels in the sensor.
In previous works~\cite{newell2016stackedhourglass, wei2016convposemachain, xiao2018simbase} based on RGB images with a CNN model, it is proved that converting the labels to a 2D heatmap and predicting the heatmap is a better way to locate the joints.
To adapt to the task based on event point cloud, we flexibly adopt a new coordinate representation, SimDR~\cite{li20212dsimdr}.
We convert the 2D labels, $(x^{\prime}, y^{\prime})$ into two 1D heat-vectors, $\boldsymbol{p_{v}}$, correspond to the size of sensor, which are one-hot and further blurred by a Gaussian core, shaped as $(H, 1)$ and $(W, 1)$, respectively: 

\begin{equation}\label{eq:8}
\left\{\begin{array}{c}
\begin{aligned}
&\boldsymbol{{p}_{v}} =\left[v_{0}, v_{1}, \ldots, v_{S}\right] \in \mathbb{R}^{S},\\
&v_{i} = \frac{1}{\sqrt{2 \pi} \sigma} \exp \left(-\frac{\left(i-v^{\prime}\right)^{2}}{2 \sigma^{2}}\right), 
\end{aligned}
\end{array}\right.
\end{equation}
where $v \in \left\{x, y\right\}$, $S \vert _{x} = W$, and $S \vert _{y} = H$.
We set $\sigma=8$ in the experiments, as the best choice for the task shown in Fig.~\ref{sigma}. 
Then we convert the value in the vectors with Min-Max Normalization, to make the largest value remain $1$. The predicted joint position $(\hat{x}, \hat{y})$ is determined by decoding the predicted vectors $\boldsymbol{\hat{p_{v}}}$ through an argmax operation to find the position of the maximum value in vectors for the $x$ axis and the $y$ axis respectively:
\begin{equation}
\boldsymbol{{pred}_{v}} =\mathop{\arg\max}\limits_{j}\left(\boldsymbol{\hat{p_{v}}}(j)\right).
\end{equation}

\subsection{3D learning model}
Our pipeline is illustrated in Fig.~\ref{backbone}.
The event point cloud is first preprocessed into the rasterized point cloud format, and then sampled to a fixed number.
The sampled rasterized event point clouds are then encoded by the backbone.
We study with three variants of the famous LiDAR point cloud processing backbones, precisely, PointNet~\cite{qi2017pointnet}, DGCNN~\cite{wang2019dgcnn}, and Point Transformer~\cite{zhao2021pointtrans}, to adapt to our mission of event point cloud based human pose estimation.
The features output from the backbone are processed by two linear layers to generate the 1D vectors.
Then, we decode the 1D vectors for x-axis and y-axis, separately, where we can obtain the predicted result of joints location.
All the three backbones can be applied to classification and segmentation in LiDAR point cloud processing, here we only modify and apply the encoder part to our task.
And the decoder is replaced by our linear layers.
For the PointNet backbone, we use a $4$-layer MLP structure and cascade event point features at multiple scales to aggregate non-local information, global features are then obtained by two symmetry functions: max pooling and average pooling. We also remove the joint alignment network~\cite{qi2017pointnet} to keep the backbone efficient, considering the natural timing order of event point cloud. As for DGCNN and Point Transformer, we keep its original structure to verify the effectiveness of the method for event point cloud processing.
Our methods can be easily deployed and integrated to work with different 3D learning architectures, adapted to event-based HPE.

\section{Experiments}

\subsection{Experiment setups}

The DHP19 dataset~\cite{calabrese2019dhp19} contains $17$ subjects.
We follow the original split setting of the dataset, which uses S1-S12 for training and test on S13-S17.
Also, we only use data from the two front camera views, same as DHP19.
We train the model using the Adam optimizer~\cite{kingma2014adam} with the Kullback–Leibler divergence loss for $30$ epochs with an initial learning rate of $1e^{-4}$ and $1e^{-5}$ for epochs $15$ to $20$, $1e^{-6}$ for epochs $20$ to $30$ on a single RTX 3090 GPU. The speed of inference is tested on an NVIDIA Jetson Xavier NX edge computing platform with a batch size of $1$, to simulate real time applications. We evaluate the results with the Mean Per Joint Position Error (MPJPE), commonly used in human pose estimation:
\begin{equation}\label{eq:11}
\mathrm{MPJPE}=\frac{1}{J} \sum_{i}^{J}\left\| pred_{i}-gt_{i} \right\|_{2}
\end{equation}
where $pred_{i}$ and $gt_{i}$ are respectively the prediction and ground truth of the $i$-th joint among $J$ joints in total for a person. The metric space of 2D error is pixels, and millimeters for 3D error.

Our model predicts 2D locations of joints directly.
To acquire the 3D locations, we project each of the 2D predictions from the pixel space to the physical space through triangulation~\cite{calabrese2019dhp19}, knowing two projection matrices as well as the positions of the two cameras. We omit the optimization on time continuity with a confidence threshold and only compare the results which are directly output.
Adapting the strategy of updating results with the output confidence may further improve the accuracy, which is not discussed in detail in this paper and we leave it for future work.

\subsection{Ablation studies}
\label{sec:ablation_studies}

\begin{table}[t]
   \caption{\textbf{Event point cloud rasterization ablations.}}
   \label{tab:rast}
   \centering
   \renewcommand\arraystretch{1.54}{\setlength{\tabcolsep}{1.20mm}{\begin{tabular}{ccccc}
            \toprule
            Input & Channel & MPJPE$_{2D}$ & MPJPE$_{3D}$   
            \\
            
            \midrule
            
            \multirow{2}{*}{Raw} 
            & $x,y,t$ & 24.75 & 310.65  \\
            & $x,y,t,p$ & 24.74 & 310.64  \\
            
            \midrule
            
            \multirow{2}{*}{Normalized} 
            & $x,y,t_{norm}$ & 7.92 & 89.62  \\
            & $x,y,t_{norm},p_{\pm1}$ & 7.61 & 86.07  \\
            
            \midrule
            
            \multirow{3}{*}{Rasterized} 
            & $x,y,t_{avg}$ & 7.77 & 87.59  \\
            & $x,y,t_{avg},p_{acc}$ & 7.40 & 84.58  \\
            & $x,y,t_{avg},p_{acc},e_{cnt}$ & \textbf{7.29} & \textbf{82.46}  \\

            \bottomrule
         \end{tabular}}}
         \vspace{-1.5em}
\end{table}
\noindent\textbf{Event point cloud rasterization.}
To analyze the contribution of the proposed rasterized point cloud format of the event, we conduct ablation studies based on PointNet with $2048$ points as input.
The results are shown in Table~\ref{tab:rast}.
With the raw data of events, it performs unsatisfactorily because the timestamps are in microseconds, which leads to a large scale difference between timestamps and x-y axes.
We then normalize the timestamps into range $[0, 1]$, and the polarity where is $0$, is changed to ${-}1$, in which the model could achieve better results.
After processing the events data into the rasterized format, the MPJPE can be further decreased, also with more proposed channels, which indicates the effectiveness of our rasterization method. We take all the five-channel representation ($x,y,t_{avg},p_{acc},e_{cnt}$) as the best choice.
In particular, the event point cloud rasterization is used as a preprocess step in this work, while it can be easily employed in the real-time processing, making use of a buffer to refresh the events information in all channels with high speed, where a preset time window length is needed.
In this way, it also has the advantage of the high temporal resolution of DVS and can perform efficiently in real-time applications.

\noindent\textbf{Number of sampling points.}
The number of points input to the model is the main reason impacting the speed of the model, while also affecting the accuracy.
We test different point numbers after sampling.
We experiment based on PointNet with the rasterized event point format.
The results are shown in Table~\ref{tab:number}.
In general, more points lead to higher accuracy and cause a decrease in speed.
As a result of adopting the sampling strategy with replacement when the number of points is not enough, the accuracy starts to decline when it reaches $7500$ points, with lots of duplicate points in the set. To hold a good balance between speed and accuracy, we chose $2048$ points for other ablation studies.

\begin{table}[t]
   \caption{\textbf{Ablations on the number of sampling points.}}
   \label{tab:number}
   \centering
   \renewcommand\arraystretch{1.625}{\setlength{\tabcolsep}{1.25mm}{\begin{tabular}{ccccc}
            \toprule
            Sampling Number & MPJPE$_{2D}$ & MPJPE$_{3D}$ & Latency (ms)   
            \\
            
            \midrule
            
            1024 & 7.49 & 85.14 & \textbf{9.43}  \\
            2048 & 7.29 & 82.46 & 12.29  \\
            4096 & \textbf{7.21} & \textbf{81.42} & 18.80  \\
            7500 & 7.24 & 81.72 & 29.18  \\

            \bottomrule
         \end{tabular}}}
        \vskip-1.0ex
\end{table}

\begin{table}[!t]
   \caption{\textbf{Comparison of different dataset labels.}}
   \label{tab:label}
   \centering
   \renewcommand\arraystretch{1.42}{\setlength{\tabcolsep}{3.2mm}{\begin{tabular}{ccc}
            \toprule
            \multirow{2}{*}{Method}&\multicolumn{2}{c}{\underline{\quad  \quad MPJPE$_{2D}$  \quad \quad}} \\
            & Last Label & Mean Label\\ 
            
            \midrule
            PointNet~\cite{qi2017pointnet}                & 7.50 & \textbf{7.29}\\
            DGCNN~\cite{wang2019dgcnn}                   & 6.96 & \textbf{6.83}\\
            Point Transformer~\cite{zhao2021pointtrans}        & 6.74 & \textbf{6.46}\\
    
            \bottomrule
         \end{tabular}}}
         \vspace{-1.5em}
\end{table}

\begin{figure}[t]
\centering
\includegraphics[width=1.0\linewidth]{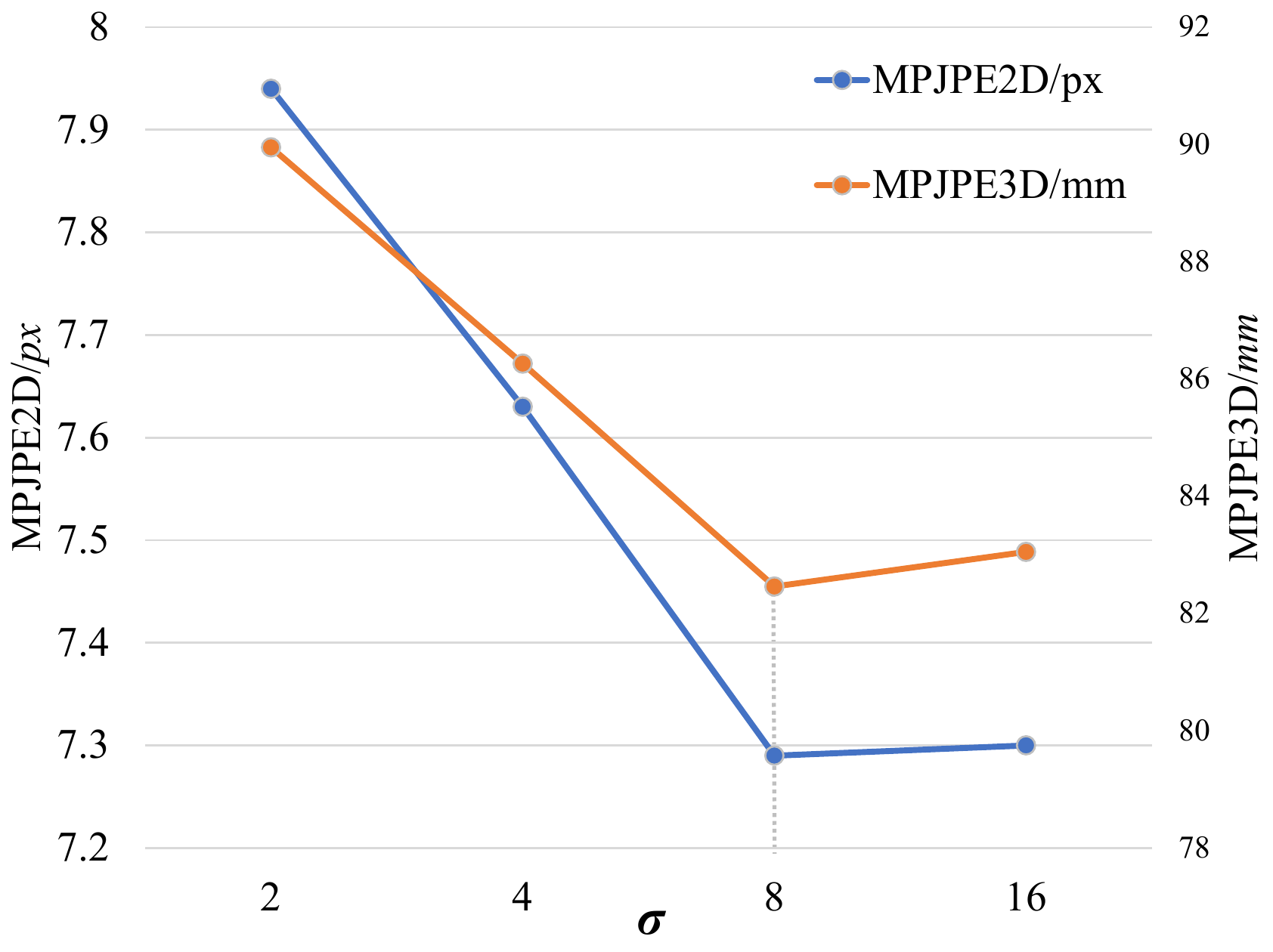}
\vskip-1ex
\caption{Parameter analysis of $\sigma$ values of label Gaussian distribution on human pose estimation accuracy. }
\label{sigma}
\vskip-2.5ex
\end{figure}

\noindent\textbf{Label of event point cloud.}
We study the parameter of $\sigma$ in Eq.~\ref{eq:8} with \emph{Mean Label} setting, and the results are shown in Fig.~\ref{sigma}. To this end, we set $\sigma=8$ as the appropriate value. As mentioned in Sec.~\ref{sec:approach_dataset}, we introduced two kinds of label setting for event point cloud based HPE.
Event point data is different from RGB frames as well as the LiDAR point clouds.
The latter two can be easily unified in time dimension and assigned an instantaneous label.
The high time resolution of event cameras leads them produce too many events in a period, while the 3D joints labels produced by Vicon motion capture system have a much lower time resolution.
We process the events in a tiny time window instead of an instant, so how to set the label properly for the task of HPE based on event streams should be considered.

Another label setting method, naming \emph{Last Label}, is explored. We process the four views separately and count $7500$ events as a window for each camera.
In this way, we assign labels nearest to the last timestamp in the window separately for each camera. We consider this method is closer to real-time applications, that is to say, the model is predicting the results closest to the current state.
Yet, this also means that the data we use from four cameras is out of sync, and could not project to 3D space with two views. A schematic diagram of the two label settings is shown in the supplementary material.

Here, we test the \emph{Last Label} setting and \emph{Mean Label} setting through all three backbones, with $2048$ points and rasterized event point format and compare MPJPE$_{2D}$ for them as shown in Table~\ref{tab:label}.
The \emph{Mean Label} performs better for the addressed task.
We consider two possible reasons.
Although the \emph{Last Label} setting seems to be more reasonable for real-time perception, the early events are too far from the label in the time dimension and we do not explicitly introduce any modules to capture the long-distance dependency, in order to maintain the simplicity of the overall structure. And the \emph{Mean Label} could take advantage of the early- and later events. On the other hand, the \emph{Mean Label} reduces single measurement error in instantaneous label recording, which is more robust for the human pose estimation task.

\begin{table*}[!t]
   \caption{\textbf{3D human pose estimation on the DHP19 dataset.}}
   \label{tab:compare_CNN}
   \centering
   \resizebox{0.95\textwidth}{!}{
   \renewcommand\arraystretch{1.435}{\setlength{\tabcolsep}{1.825mm}{\begin{tabular}{ccccccc}
            \toprule
            Input & Method & MPJPE$_{2D}$ & MPJPE$_{3D}$ & $\#$Params (M) & $\#$MACs (G) & Latency (ms)    
            \\
            
            \midrule
            
            \multirow{5}{*}{2D Event Frames} 
            & Pose-ResNet18†~\cite{xiao2018simbase} & 5.37 & 61.03 & 15.4 & 8.30 & 68.07 \\
            & Pose-ResNet50†~\cite{xiao2018simbase} & \textbf{5.28} & \textbf{59.83} & 34.0 & 12.91 & 93.56 \\
            & MobileHP-S†~\cite{choi2021mobilehumanpose} & 5.65 & 64.14 & 1.83 & 0.69 & 48.09 \\
            & LeViT-128S†~\cite{graham2021levit} & 7.68 & 87.79 & 7.87 & \textbf{0.20} & \textbf{14.40} \\
            & DHP19~\cite{calabrese2019dhp19} & 7.67 & 87.90 & \textbf{0.22} & 3.51 & 27.55 \\
            
            \midrule
            
            \multirow{3}{*}{3D Event Point Cloud}
            & PointNet~\cite{qi2017pointnet} & 7.29 & 82.46 & 4.46 & \textbf{1.19} & \textbf{12.29} \\
            & DGCNN~\cite{wang2019dgcnn} & 6.83 & 77.32 & 4.51 & 4.91 & 127.96 \\
            & Point Transformer~\cite{zhao2021pointtrans} & \textbf{6.46} & \textbf{73.37} & \textbf{3.65} & 5.03 & 497.27 \\

            \bottomrule
         \end{tabular}}}}
        \vskip-1ex
\end{table*}

\subsection{Comparison on the DHP19 dataset}
Our methods can be easily integrated with and consistently effective for different 3D learning architectures to attain robust event-based human pose estimation. Table~\ref{tab:compare_CNN} shows the results of the proposed method with $2048$ points as input versus the 2D event-frame-based methods († indicates our reimplementation).
We follow the Simple Baseline~\cite{xiao2018simbase} to train the models of Pose-ResNet18 and Pose-ResNet50 with constant count event frames.
In addition, we implement a model based on LeViT~\cite{graham2021levit}, an advanced backbone with faster inference. MobileHumanPose~\cite{choi2021mobilehumanpose} is also tested as another backbone for 2D prediction with the same framework as ours.
CNN models with much more parameters and computation overheads perform better in this scene, which is reasonable.
Our PointNet is superior to DHP19 in terms of both accuracy and speed, which is demonstrated to be very effective for the dedicated task.
DGCNN and Point Transformer achieve higher accuracy, even close to the results of Pose-ResNet18 and Pose-ResNet50.

For this movement-driven task demonstrated in DHP19~\cite{calabrese2019dhp19}, we count every piece of data in the test dataset and set the minimum time span value of about $36ms$ as the inference latency that the model should achieve to meet the real-time prediction requirement for event cameras in this scene.
The latency \textit{vs.} MPJPE$_{3D}$ for different models are illustrated in Fig.~\ref{intro}, where our method achieves fast inference with good performance. Our PointNet with $2048$ points only has a latency of $12.29ms$, which is ideally suitable for efficiency-critical mobile applications.

\subsection{Results visualization and real-world estimation}
\begin{figure*}[t]
\centering
\includegraphics[width=0.86\linewidth]{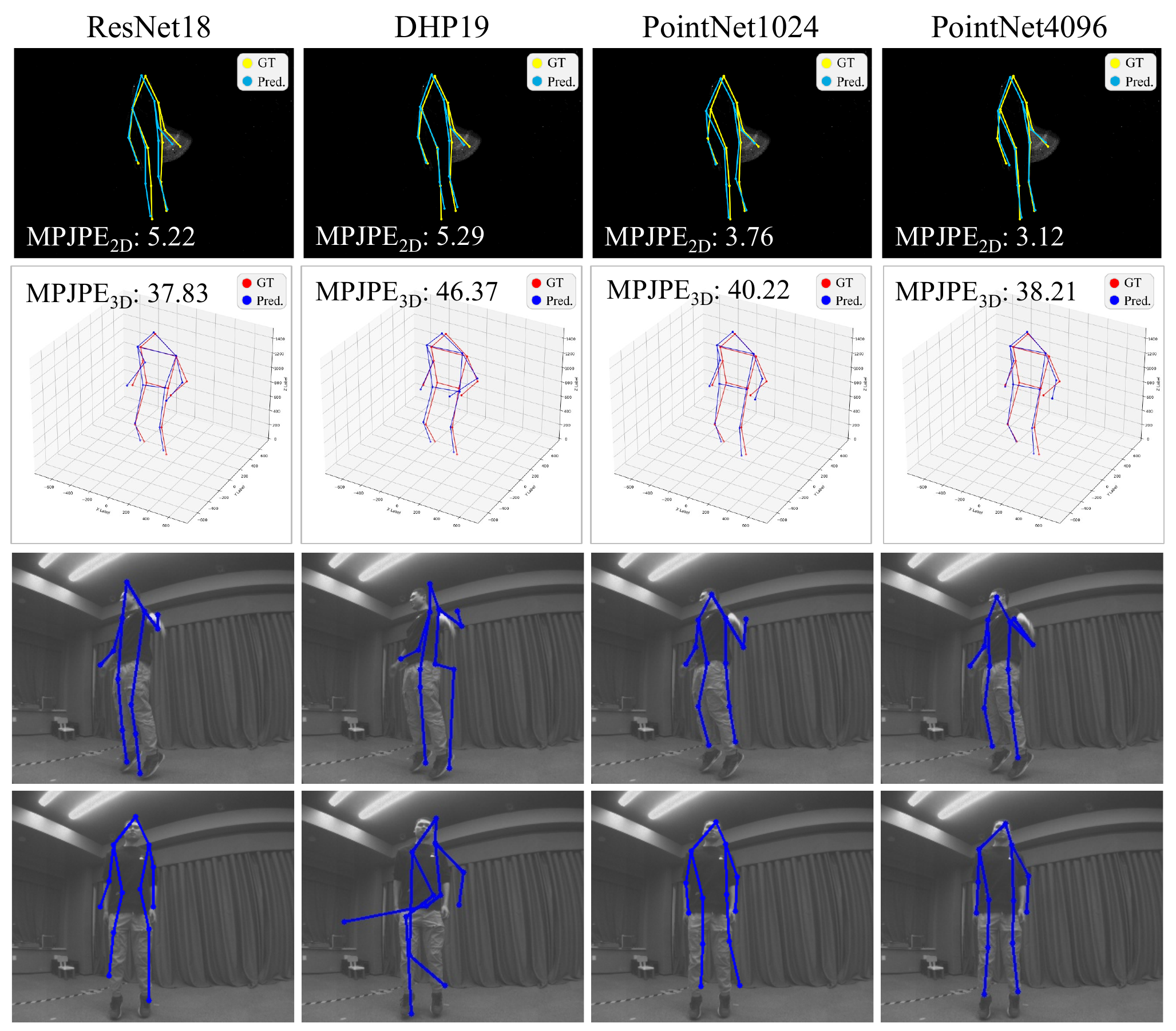}
\vskip-1ex
\caption{Results visualization for different models. The first row shows the results in 2D (yellow for ground truth, blue for prediction). Our method achieves a more successful prediction for our PointNet even with many static limbs. The second row shows the results in 3D (red for ground truth, blue for prediction). The last two rows show the results on our own device, where the DHP19 model fails.}
\label{compare}
\vskip-3ex
\end{figure*}

Fig.~\ref{compare} visualizes the results of different models. We plot the 2D joints together with skeletons over the constant count event frames, while the results for event point cloud based methods are predicted from point sets.
To explore the generalizability of the proposed method,
we also test different models in real-world scenarios by capturing event data with a DAVIS camera, which has a resolution of $262{\times}320$. Although DHP19 is also recorded by the DAVIS camera, a kind of event camera that transmits DVS events in addition to APS (Active Pixel Sensor) frames, only the DVS output was recorded and the APS output was discarded due to bandwidth constraints.
To make use of the point-wise methods, events are scaled to the DHP19 resolutions ({$260{\times}346$}), and the results are finally re-scaled to the resolution of the device itself.
Results are shown over the APS frames of our DAVIS, with no ground truth.

It can be clearly seen that for the new data with different devices, 
our approach still delivers impressive results,
which reveals that our human pose estimation models generalize well to such unseen domains. However, the real-time CNN method DHP19 fails on the entire sequence, where we use the original open source pre-training model~\cite{calabrese2019dhp19} (see supplementary material for details).
Moreover, we find that the DHP19 model fails on the new data despite the conducted denoising and filtering, which may be caused by the small amount of parameters leading to poor generalization to a new device as well as different views. Note that the input to our method is the raw event point cloud, which we believe is important for efficient event HPE, considering that the denoising process will introduce additional computation.

\section{Conclusion}
In this work, we have explored the potential of efficient Human Pose Estimation (HPE) by point-wise networks through directly processing the event points. We proposed the rasterized event point cloud, a new event representation, which could maintain the 3D features from multiple statistical cues.
Experiments based on three famous point-wise backbones are carried out, attaining different trade-offs between speed and accuracy.
In particular, Point Transformer reaches the best performance, while PointNet is the fastest.
We have also discussed the label setting for the new perspective of the mission, and the \emph{Mean Label} is more productive.
We expect this work to pave the way for subsequent event point cloud HPE, and we will continue to explore the potential of event point cloud based efficient pipelines in other fields.

\noindent\textbf{Acknowledgments.}
This work was supported in part by Sunny Optical Technology (Group) Co. Ltd., in part by the National Natural Science Foundation of China (NSFC) under Grant No. 12174341, and in part by the Federal Ministry of Labor and Social Affairs (BMAS) through the AccessibleMaps project under Grant 01KM151112.

\clearpage
{\small
\bibliographystyle{ieee_fullname}
\bibliography{egbib}
}

\newpage
\appendix
\section{Label setting}

\begin{figure*}[t]
\centering
\includegraphics[width=1.0\linewidth]{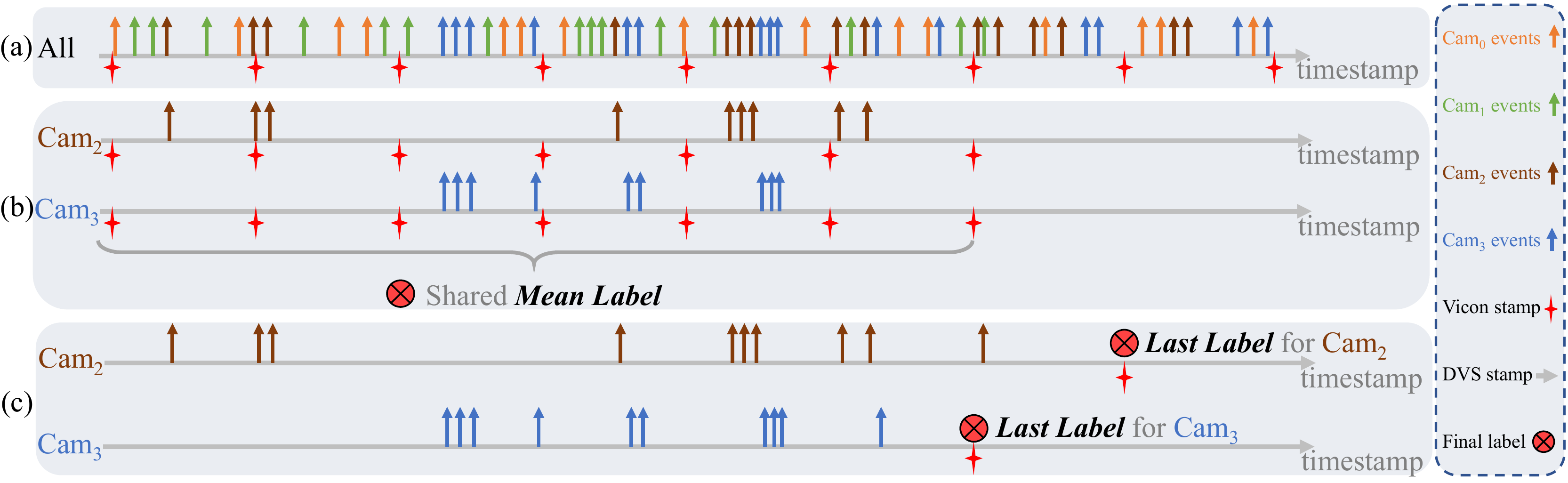}
\caption{Schematic diagram of label setting for event point cloud based human pose estimation. (a) The event point cloud output frequency of all four event cameras and the acquisition frequency of the motion capture device Vicon. (b) Determine the shared mean label for event camera 2 and event camera 3. (c) Determine the last label for event camera 2 and event camera 3, respectively. }
\label{label}
\end{figure*}

In DHP19~\cite{calabrese2019dhp19}, the event streams from the four event cameras are merged and have an ordered monotonic timestamp. In Fig.~\ref{label}(a), an amount of $60$ events, $15$ for each camera with various colors, in an ordered timestamp are shown schematically.
And the time resolution of the output label from Vicon system is much lower than that of the event camera, which is shown as the red star marker on the timestamp axis for illustration. The \emph{Mean Label} setting is conducted under a fixed number of events for all cameras in total, \textit{e.g.,} $40$ events. By counting all events, a shared \emph{Mean Label} is generated, which is the mean value of all labels in the window, shown in Fig.~\ref{label}(b). For the \emph{Last Label} setting, when a fixed number of events is counted for each camera, \textit{e.g.,} $10$ events, the label nearest the last event is set as the label for each camera separately, shown in Fig.~\ref{label}(c).

As it can be seen, the number of events for different cameras may differ in \emph{Mean Label} setting. Also, there may be a difference with the time span of events in \emph{Last Label} setting for different cameras.  The \emph{Last Label} setting leads the model to predict the results closest to the current state, while the \emph{Mean Label} taking advantage of the early- and later events to predict results corresponding to the interval. We have studied the performance of the two different label settings in the experiment section and the \emph{Mean Label} performs better for the addressed task.

\section{Event point numbers in the dataset}
When generating event point cloud data for the \emph{Mean Label} setting, events are accumulated together for the four views, and some cameras may not be calculated, leading to an empty point cloud or have too few points.
It also occurs when generating event frames which correspond to an empty image.
We consider that this is a meaningful finding when using the DHP19 dataset with the \emph{Mean Label} setting, as it is ignored in frame-based methods where empty images exist in the dataset. Compared with the previous event-frame-based DHP19~\cite{calabrese2019dhp19}, we remove the data with points fewer than $1024$ when training our model, and the testing set is the same for both with all data. Also, we experiment a training with all data for PointNet-2048. It causes a bit accuracy decrease, but still better than DHP19.
the result is shown in Tab.~\ref{tab:1024_data}.
\begin{table}[h]
\renewcommand\thetable{B}
\footnotesize
   \centering
   \resizebox{0.45\textwidth}{!}{
   \renewcommand\arraystretch{1.0}{\setlength{\tabcolsep}{1.5mm}{\begin{tabular}{cccc}
            \toprule
            Method & $N < 1024$ samples & MPJPE$_{2D}$ & MPJPE$_{3D}$\\
            
            \midrule
            
            DHP19~\cite{calabrese2019dhp19} & \textit{with} & 7.67 & 87.90 \\
            PointNet-2048 & \textit{with} & 7.40 & 83.93 \\
            PointNet-2048 & \textit{w/o} & 7.29 & 82.46 \\
            
            \bottomrule
         \end{tabular}}}}
         \caption{Ablation experiment on training data.}
\label{tab:1024_data}
\end{table}

\section{Time slice $K$}
We conduct ablations on time slice $K$
(see Tab.~\ref{tab:K_ablation}).
When $K$ is smaller, the information density of the rasterized event point cloud is higher, but the temporal resolution will be reduced, thus the choice needs to be weighed. We note the HPE task is not sensitive to $K$. Our model achieves satisfactory performance at $K{=}4$, consistent with our setting.

While the event point cloud rasterization setting works well in our task of HPE, it does not have to be the optimal choice when transferring to other event-driven vision applications. We will further explore in this direction like the setting of $K$ in more event-based learning problems such as optical flow estimation and human pose tracking.

\begin{table}[h]
\renewcommand\thetable{A}
\footnotesize
   \resizebox{0.45\textwidth}{!}{
   \renewcommand\arraystretch{1.0}{\setlength{\tabcolsep}{2mm}{\begin{tabular}{ccccc}
            \toprule
            \multirow{2}{*}{$K$} & \multicolumn{2}{c}{\underline{\quad PointNet-2048 \quad}} & \multicolumn{2}{c}{\underline{\quad PointNet-4096 \quad}}\\
            & MPJPE$_{2D}$ & MPJPE$_{3D}$ & MPJPE$_{2D}$ & MPJPE$_{3D}$ \\
            \midrule
            1 & 7.29 & 82.40 & 7.24 & 81.74 \\
            2 & \textbf{7.28} & 82.49 & 7.23 & 81.99 \\
            4 & 7.29 & \textbf{82.46} & \textbf{7.21} & \textbf{81.42} \\
            8 & 7.31 & 83.18 & 7.23 & 81.67 \\
            
            \bottomrule
         \end{tabular}}}}
         \caption{Ablation experiment on $K$.}
\label{tab:K_ablation}
\end{table}

\section{More visualizations}
\begin{figure*}[b]
\centering
\includegraphics[width=1.0\linewidth]{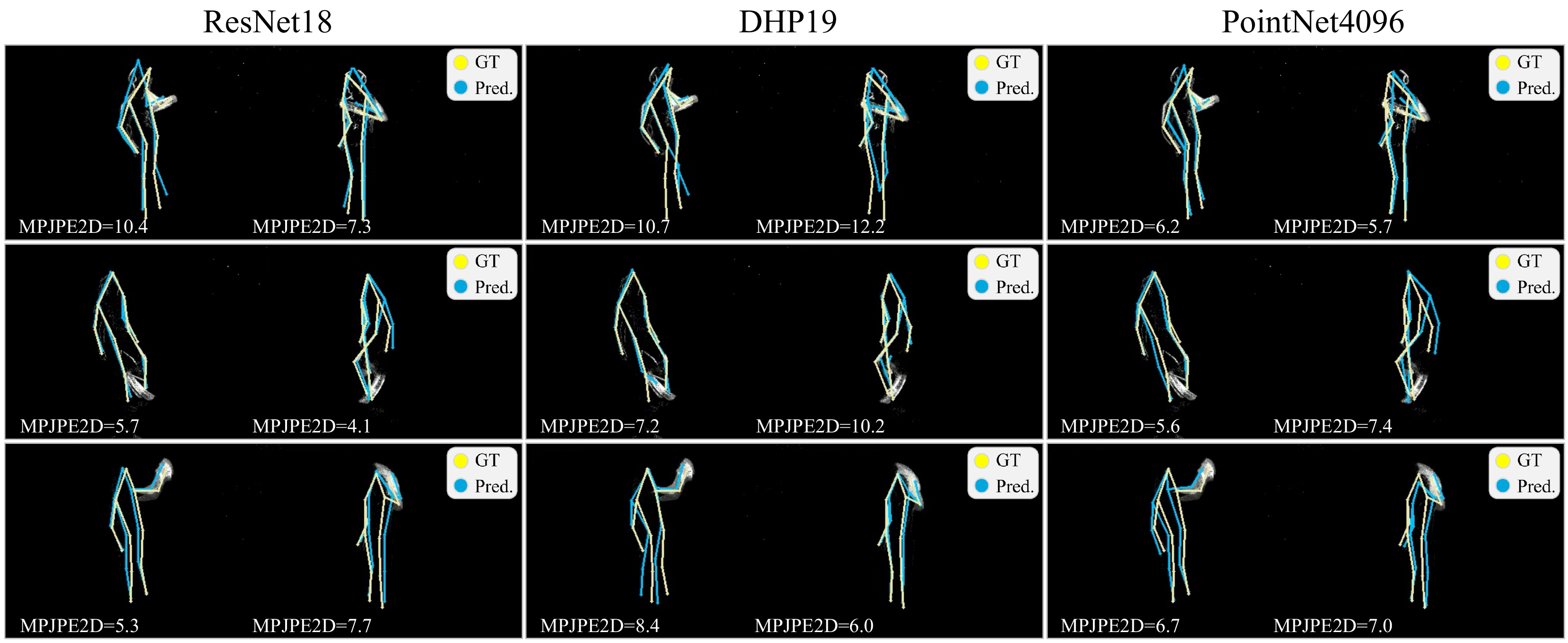}
\caption{2D results visualization with two front views on the DHP19 test dataset for different models (yellow for ground truth, blue for prediction). The three movements from top to bottom are ``Punch straight forward left'', ``Kick forwards left'', and ``Figure-8 left hand''.}
\label{2d}
\end{figure*}

\begin{figure*}[b]
\centering
\includegraphics[width=1.0\linewidth]{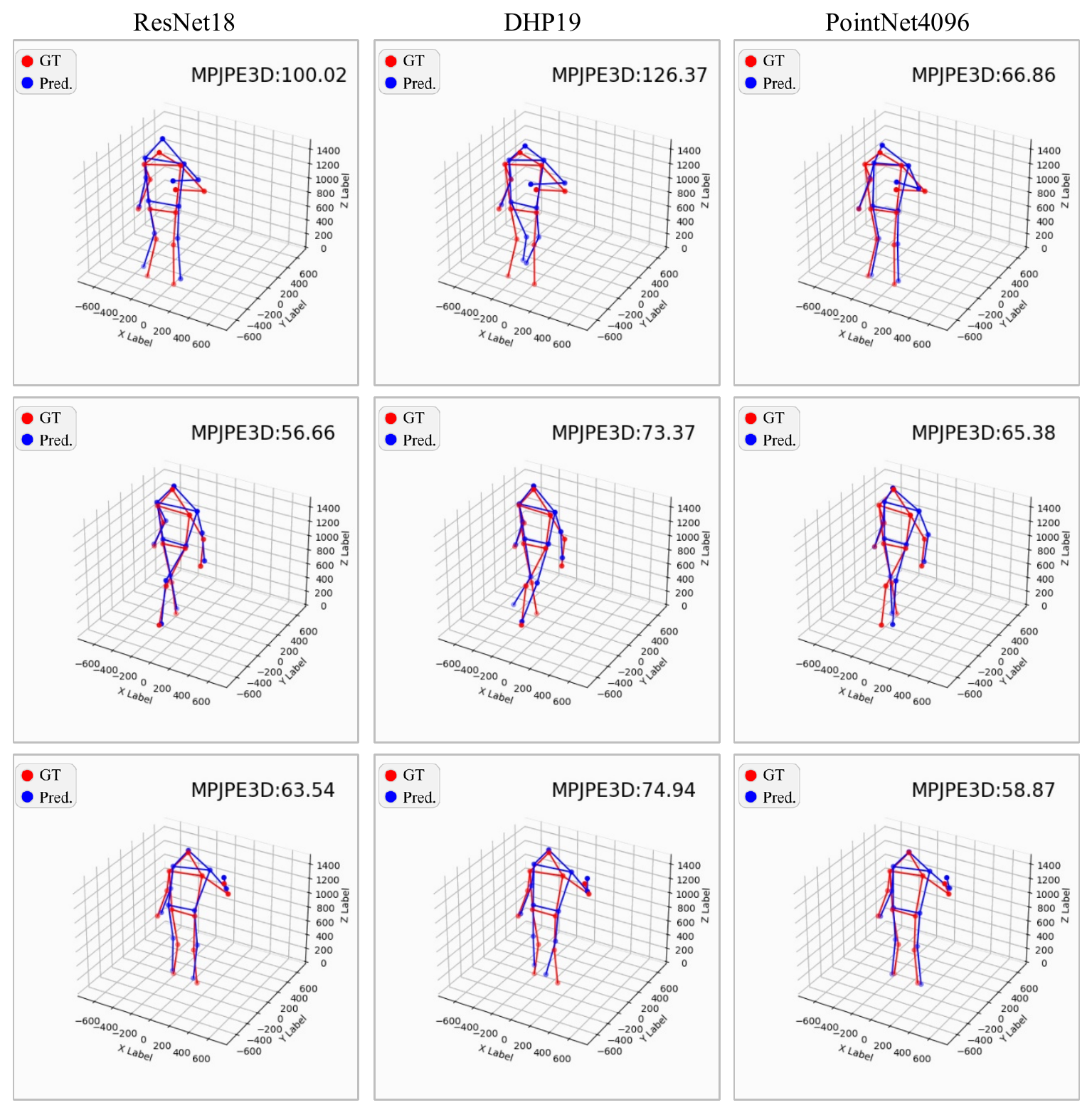}
\caption{3D results visualization on the DHP19 test dataset for different models (red for ground truth, blue for prediction). The three movements from top to bottom are ``Punch straight forward left'', ``Kick forwards left'', and ``Figure-8 left hand''.}
\label{3d}
\end{figure*}

Fig.~\ref{2d} and Fig.~\ref{3d} display more results on the DHP19 test dataset~\cite{calabrese2019dhp19} for different models, which are selected from movements ``Punch straight forward left'', ``Kick forwards left'', and ``Figure-8 left hand''. 
The 2D results show two front views, which are used to project the predictions from the pixel space to the physical space through triangulation. 
We find that our event point cloud based method is more robust when facing static limbs than the DHP19 model. Static limbs during the movement, such as legs in movement ``Figure-8 left hand'', generate few events which lead to invisible parts when accumulating events to a constant count event frame. However, such few events are retained in the event point cloud, and they could be processed by the point-wise backbone more effectively, which we think is the main reason for the superiority of our method in predicting keypoints for static limbs. 

Also, in the unseen domains, such as different views and noise with our own device, the event point cloud with a multidimensional feature outperforms the event-frame-based DHP19 and is robust to the noise brought by the background as well as the device itself. 
A video is submitted as a supplementary with more continuous visualizations, which further demonstrates the high efficiency of our event point cloud based human pose estimation framework. 

\end{document}